\documentclass[letterpaper, 10 pt, conference]{ieeeconf}  

\IEEEoverridecommandlockouts                              

\usepackage{cite}
\usepackage{amsmath,amssymb,amsfonts}
\usepackage{algorithmic}
\usepackage{graphicx}
\usepackage{textcomp}
\usepackage{xcolor}
\usepackage{amsfonts}
\usepackage{amsmath}
\usepackage{subfigure}
\usepackage{diagbox}
\usepackage{tikz}
\usepackage{booktabs}
\usepackage[ruled,vlined,linesnumbered]{algorithm2e}

\title{\LARGE \bf
Digital Twin Synchronization: Bridging the Sim-RL Agent to a Real-Time Robotic Additive Manufacturing Control
}

\author{Matsive Ali\textsuperscript{†}, Sandesh Giri\textsuperscript{†}, Sen Liu\textsuperscript{†}*, and Qin Yang*
\thanks{\textsuperscript{†}Department of Mechanical Engineering, University of Louisiana at Lafayette, Lafayette, LA 70503, USA, *Corresponding author: email:{\tt\small sen.liu@louisiana.edu}}%
\thanks{*Corresponding author: Computer Science \& Information Systems Department, Bradley University, Peoria, IL 61625, USA, email: 
        {\tt\small is3rlab@gmail.com}}%
}

\begin{document}

\maketitle
\thispagestyle{empty}
\pagestyle{empty}

\begin{abstract}

With the rapid development of deep reinforcement learning technology, it gradually demonstrates excellent potential and is becoming the most promising solution in the robotics. 
However, in the smart manufacturing domain, there is still not too much research involved in dynamic adaptive control mechanisms optimizing complex processes.
This research advances the integration of Soft Actor-Critic (SAC) with digital twins for industrial robotics applications, providing a framework for enhanced adaptive real-time control for smart additive manufacturing processing. The system architecture combines Unity's simulation environment with ROS2 for seamless digital twin synchronization, while leveraging transfer learning to efficiently adapt trained models across tasks. 
We demonstrate our methodology using a Viper X300s robot arm with the proposed hierarchical reward structure to address the common reinforcement learning challenges in two distinct control scenarios. 
The results show rapid policy convergence and robust task execution in both simulated and physical environments demonstrating the effectiveness of our approach. 

\end{abstract}

\section{Introduction}
\label{main}


As a significant aspect of artificial intelligence, reinforcement learning (RL) provides an approach to helping robots develop self-learning capability. RL can increase the adaptability of robotics systems, which is an important feature in order to deal with a complex and dynamic environment \cite{polydoros2017survey,yang2024bayesian}. However, there are many challenges to implementing RL in the robotic domain due to hardware limitations (such as mechanical system design and computer capabilities) and software restrictions (like algorithm efficiency).
Reinforcement learning (RL) has emerged as a promising approach for developing such adaptive control systems, offering advantages over traditional control methods through its ability to optimize behavior through environmental interaction \cite{kulkarni2012reinforcement}. 
Unlike supervised learning approaches that rely on labeled datasets, RL enables autonomous exploration and strategy refinement, making it particularly valuable for robotics applications where continuous adaptation is essential.

From the manufacturing perspective, the challenges include sample inefficiency during training, computational intensity, and potential instability in learned policies \cite{thompson2019reinforcement}. Additionally, the direct application of RL algorithms in physical manufacturing systems risks equipment damage and production disruption during the learning phase.
Recent applications have demonstrated RL's potential across various manufacturing domains. Wang et al. \cite{wang2018deep} successfully applied RL to optimize CNC machining parameters in real-time, achieving improvements in surface quality while reducing tool wear. In the automotive sector, Loffredo et al. \cite{loffredo2023reinforcement} implemented RL-based energy-efficient control systems for multi-stage production lines, effectively balancing energy consumption with throughput requirements. 

\begin{figure}[t]
    \centering
    \includegraphics[width=1\linewidth]{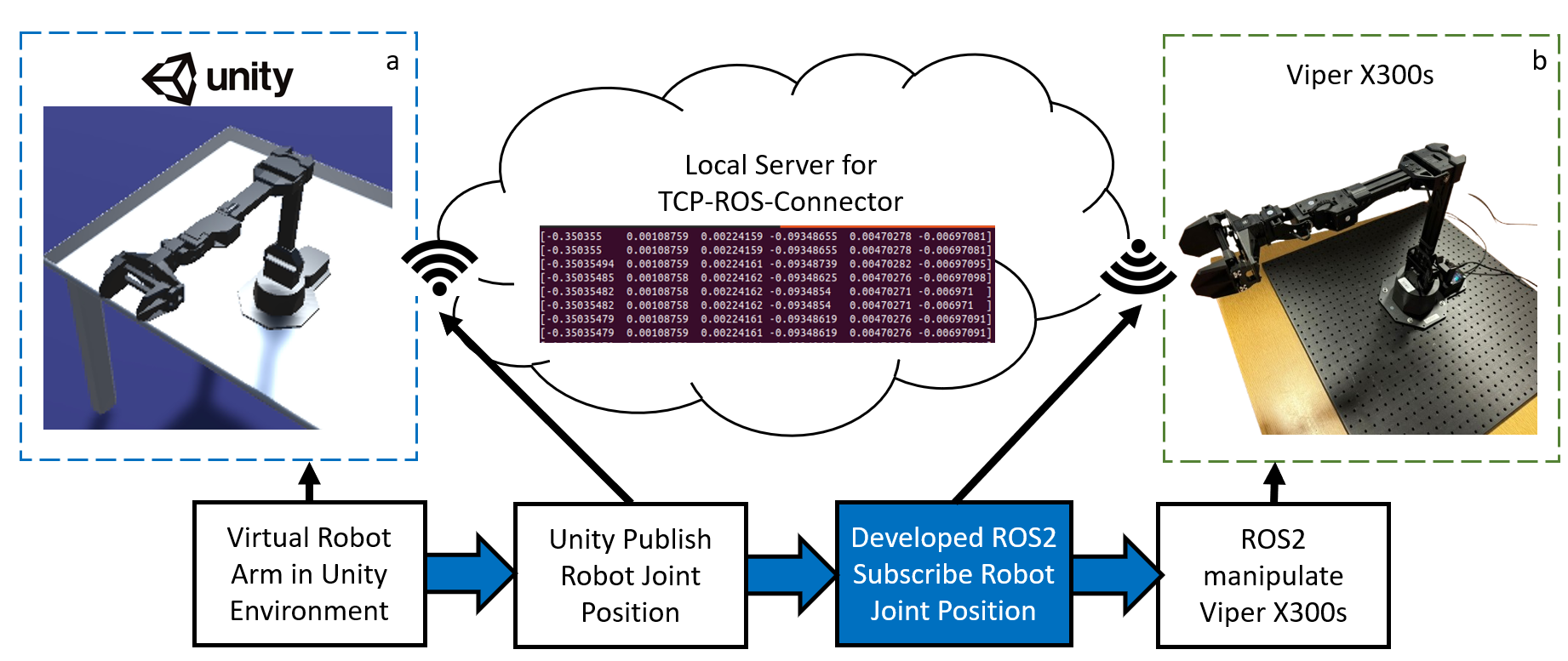}
    \caption{Digital Twin Synchronization process of (a) virtual robot in Unity and (b) the physical robot Viper X300s.}
    \label{fig:DT-Fig1}
\end{figure}

However, robotic arm control and path optimization in industrial manufacturing rely on predefined models and static decision-making processes. Especially in path planning, robots must calculate the most efficient trajectory while avoiding obstacles, complying with kinematic constraints, and reacting to real-time environmental feedback, which makes path optimization a complex and time-consuming task. 

To address this gap, we propose a digital twin synchronization method that transforms the well-trained simulation RL (sim-RL) agent through Unity into a real robot arm in real time within the Unity and ROS2 framework. Moreover, to overcome RL challenges such as slow convergence, local minima, and instability in the training process, we introduce the Hierarchical Reward Structure (HRS) integrating with the Soft Actor-Critic (SAC) \cite{haarnoja2018soft} method to achieve sample efficiency. The main contributions of this research is listed as below.
\begin{itemize}
    \item \textbf{Integration of SAC with Real-Time Synchronization:} Combine the SAC RL algorithm with real-time synchronization for controlling the Viper X300s robot arm in both virtual and physical environments.
    \item \textbf{Task Scenarios:} Explore two distinct robotic tasks: a static target-reaching task and a dynamic goal-following task to evaluate the RL agent's performance for line scanning settings.
    \item \textbf{Hierarchical Reward Structure (HRS):} Leverage HRS techniques to enhance model adaptation across tasks, accelerating training efficiency and convergence. Develop reward mechanisms tailored to overcome RL-specific challenges such as local minima and instability in learning.
    \item \textbf{Comprehensive Evaluation of Performance Metrics:} Assess cumulative reward, episode length, policy loss, and value prediction accuracy in both virtual and physical settings.
\end{itemize}

\section{Methodology}

\subsection{Experimental Setup and Digital Twin Synchronization}
Our experimental setup utilizes the Viper X300s, a Six Degree of Freedom (DOF) robot arm designed by Trossen Robotics for research applications. With a span of 1500mm and a payload capacity of 750g, the Viper X300s offers researchers a compact yet powerful platform for experimenting with robotic functionalities. Through ROS inter-bridging, researchers can simulate tasks in Unity’s virtual environment and then validate them on the Viper X300s, creating a seamless feedback loop between virtual testing and physical deployment. We built a ROS2 package to receive data from the virtual model of the Viper X300s robot arm in Unity through a local TCP server. Fig. \ref{fig:DT-Fig1} illustrates the establishing digital twin connection. 
The Unity ROS-TCP-Connector extension enables seamless publishing of ROS2 messages to a local cloud server, allowing bidirectional communication between Unity and ROS-based systems. These messages are subscribed by ROS2 for processing on the physical Viper X300s robot arm, establishing real-time digital twin synchronization. This synchronization has a latency of only about 20 milliseconds, ensuring near-instantaneous reflection of virtual movements on the physical counterpart, which allows for testing and validation of robotic tasks in a simulated environment before deployment \cite{plasberg2022towards}. 

\subsection{Reinforcement Learning in Unity}

\subsubsection{Unity ML-Agents Framework for Reinforcement Learning}

\begin{figure}[t]
    \centering
    \includegraphics[width=1\linewidth]{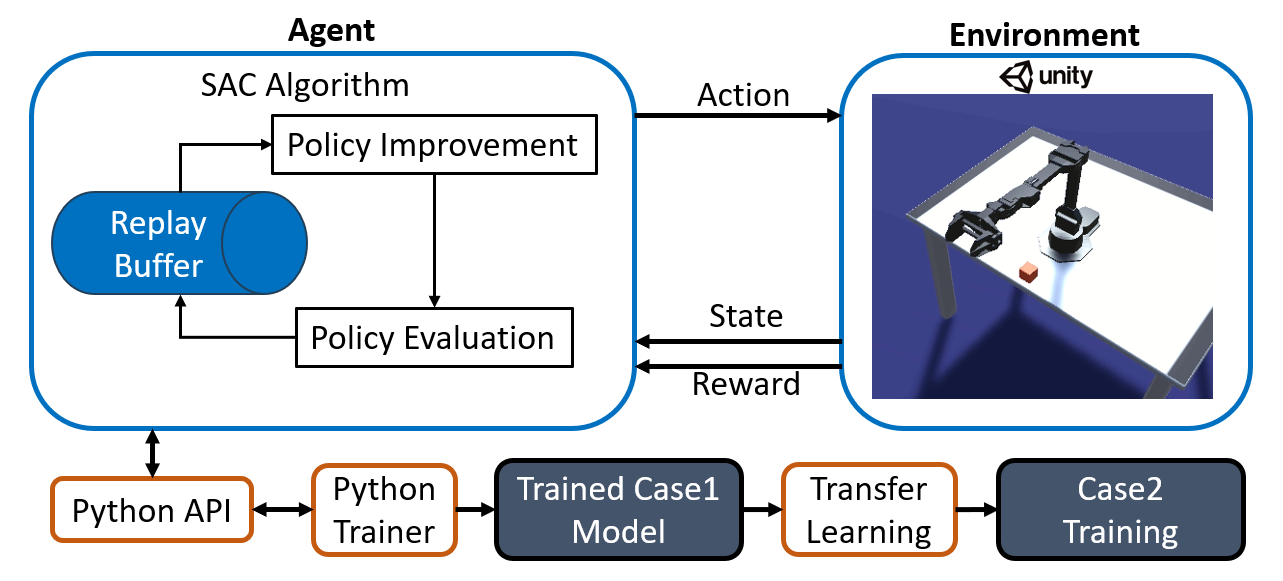}
    \caption{Training Process of Soft Actor-Critic (SAC) Reinforcement Learning Algorithm in Unity for Virtual Viper X300s Robot Arm for Case 1 and hierarchical reward training model transfer to Case 2.}
    \label{fig:Fig2}
\end{figure}

\begin{figure}[t]
    \centering
    \includegraphics[width=1\linewidth]{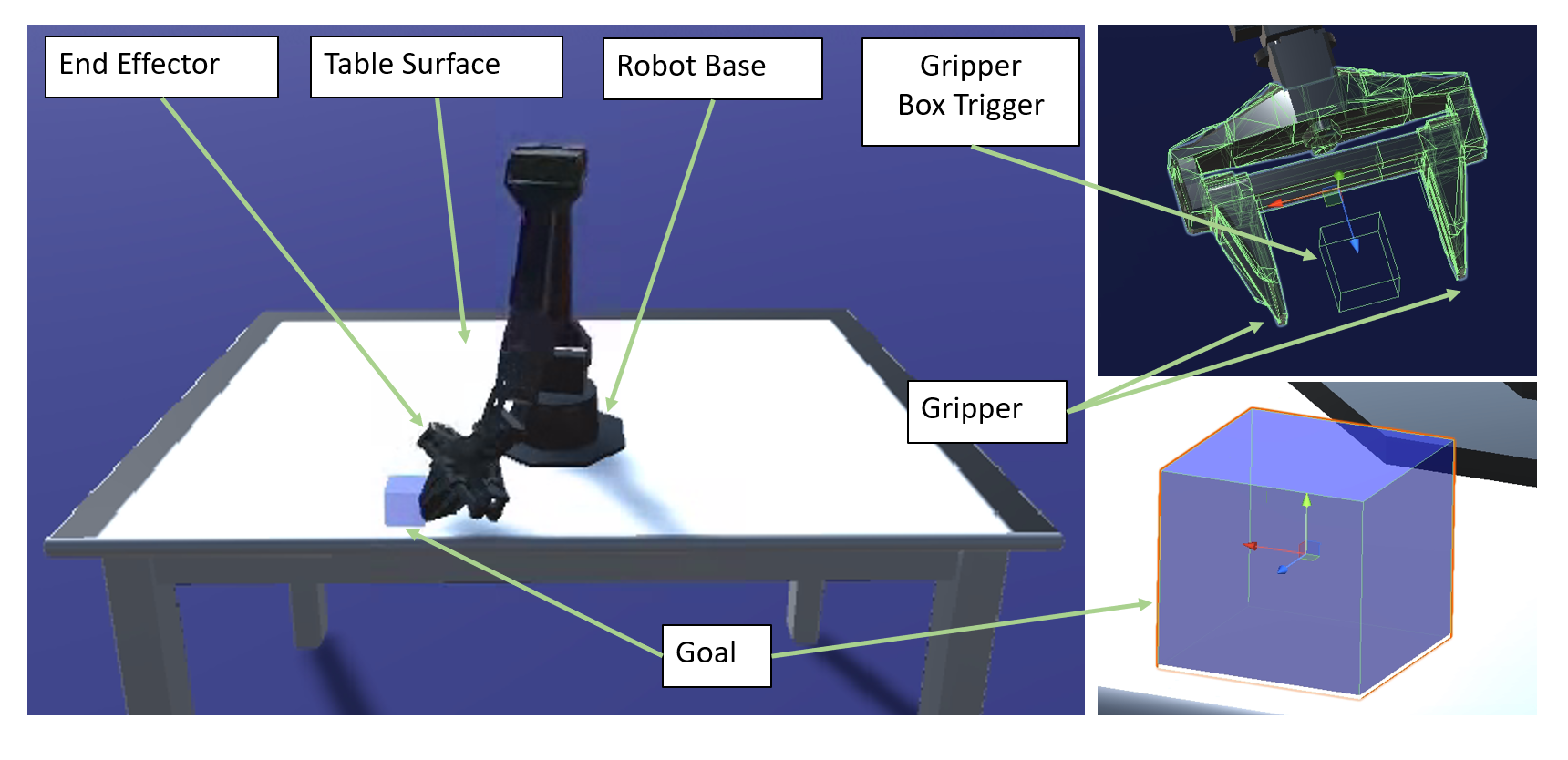}
    \caption{Unity environment with Virtual Robot Arm and goal.}
    \label{fig:fig3}
\end{figure}

Our established algorithms environment was run in Unity \cite{engelbrecht2023unity} consisting of different objects and a virtual robot arm as illustrated in Fig. \ref{fig:Fig2}. The algorithms were trained through Python Trainer. Initially, the states of the environment were fed to the RL agent so that it could gain an understanding of the working environment. After getting the state, the agent took a set of action policies to assess the change in state and the reward. These changes were further fed into the agent to store the action policies in the replay buffer. Following this implementation, the agent improved its set of action policies and then evaluated them. After that, the environment communicated the state and reward back to the agent. This iteration was done throughout the training, and after the Case 1 training was completed, the trained model was transferred to the Case 2 training, which also utilized the trained models' weights to adjust and improve its learning. The training process has been shown in Fig. \ref{fig:fig3}.

\subsubsection{Action Space for Robot Arm Control}
To control the robot arm, we defined a discrete action space with seven action branches: six for each joint’s movement and one for gripper translation, as outlined in Table \ref{tab:Discrete Action Space}. Each action branch has three options (e.g., turn left, stop, turn right), allowing fine-grained control over the robot’s joint movements. The agent receives state information from Unity that includes positional data for the gripper, goal position, end-effector, and distance to the goal, all of which are used to help the agent learn and optimize its movements for task completion. 


\begin{table}[t]
\caption{Discrete Action Space for Agent Controlling Robot Arm in Unity}
\begin{tabular*}{\hsize}{@{\extracolsep{\fill}}lccc@{}}
\hline
Discrete Action \\ Space Brach & Controls & Brach Size & Brach Actions \\
\hline
1 & Joint 1 & 3 & Turn Left, Stop, Turn Right \\
2 & Joint 2 & 3 & Turn Left, Stop, Turn Right \\
3 & Joint 3 & 3 & Turn Left, Stop, Turn Right \\
4 & Joint 4 & 3 & Turn Left, Stop, Turn Right \\
5 & Joint 5 & 3 & Turn Left, Stop, Turn Right \\
6 & Joint 6 & 3 & Turn Left, Stop, Turn Right \\
7 & Gripper & 3 & Open, Stop, Close \\
\hline
\label{tab:Discrete Action Space}
\end{tabular*}
\end{table}

\subsubsection{Hierarchical Reward Structure Policy}
We design a hierarchical reward structure policy to improve learning efficiency. 
The tables \ref{tab:Case1}, \ref{tab:Case2} and \ref{tab:Case3} outline the reward framework designed for two cases in the reinforcement learning task, highlighting the primary goals, sub-goals, reward values, and the utilization of transfer learning to enhance model adaptation. The listed framework strategically combines positive reinforcement to achieve objectives and maintain stability with penalties for undesirable actions, while leveraging transfer learning from simpler tasks to more complex ones.
This strategy decomposes the task into sub-goals, each with its own reward structure, allowing the agent to achieve incremental successes and maintain consistent learning progress. By incorporating intermediate rewards for sub-goals, the agent learns high-level strategies and low-level actions more effectively, improving training scalability and stability. 
To further accelerate learning, we used transfer learning based on the hierarchical reward structure, where the trained model from Case 1 was fine-tuned for Case 2, leveraging pre-learned features to reduce training time and computational costs. Transfer learning enhances performance and efficiency, particularly in scenarios where the new task shares similarities with a previously trained model.

\begin{table}[t]
\caption{Robot Learning Case 1: Touch Goal with Grippers Box Trigger}
\begin{tabular*}{\hsize}{@{\extracolsep{\fill}}p{2cm} p{2cm} c p{3cm}@{}}
\toprule
Goal & Sub Goals & Rewards & Transfer Learning \\
\hline
{\parbox{3cm}{Touch Goal with \\ Grippers Box \\ Trigger}} & 
Touch Goal with Grippers & +1 & None \\
& Keep Robot Arm Upright & 0 to +3 & \\
& Touch Table with Gripper & -1 & \\
& Gripper goes below Table Surface & -1 & \\
& Gripper goes behind Robot base & -1 & \\
& Per 300 steps agent uses & -1 & \\
\hline
\label{tab:Case1}
\end{tabular*}
\end{table}
\begin{table}[t]
\caption{Robot Learning Case 2: Follow Linearly Moving Goal}
\begin{tabular*}{\hsize}{@{\extracolsep{\fill}}p{2cm} p{2cm} c p{3cm}@{}}
\toprule
Goal & Sub Goals & Rewards & Transfer Learning \\
\hline
\centering{\parbox{3cm}{Follow Linearly \\ Moving \\ Goal}} & 
Touch Goal with Grippers & +10 & Using Case 1 \\
& Keep Robot Arm Upright & 0 to +3 & \\
& Touch Table with Gripper & -1 & \\
& Gripper goes below Table Surface & -10 & \\
& Gripper goes behind Robot base & -10 & \\
\hline
\label{tab:Case2}
\end{tabular*}
\end{table}
\begin{table}[t]
\caption{Robot Learning Case 3: Case 2 without Transfer Learning}
\begin{tabular*}{\hsize}{@{\extracolsep{\fill}}p{2cm} p{2cm} c p{3cm}@{}}
\toprule
Goal & Sub Goals & Rewards & Transfer Learning \\
\hline
{\parbox{3cm}{Follow Linearly \\ Moving \\ Goal}} & 
Touch Goal with Grippers & +10 & None \\
& Keep Robot Arm Upright & 0 to +3 & \\
& Touch Table with Gripper & -1 & \\
& Gripper goes below Table Surface & -10 & \\
& Gripper goes behind Robot base & -10 & \\
\hline
\label{tab:Case3}
\end{tabular*}
\end{table}

In the first case (Case 1) of research, the robot arm was initially trained to touch a goal object. As robot arm should not be in any slouching posture, a lower-level reward policy was added to make sure that the robot always maintained an upright posture. Then, a higher-level reward policy was assigned to the agent if the gripper box trigger collided with the goal. When the collision occurred, it triggered the end of the episode, and the goal position gets reassigned to a random position nearby. After successfully training the model of Case 1, transfer learning was applied to train the second case (Case 2) related to linear scanning with the reward policy mentioned in Table \ref{tab:Case2}. This made the trainer use the previous models’ neural networks weights to converge with more efficiency. The goal for Case 2 was then assigned to keep robot arm upright while following the goal as it moved linearly in the X-direction. In order to show the effectiveness of the hierarchical structure training of case 2, the third case (Case 3) was designed with the same objectives as Case 2 but without any transfer learning from Case 1. The reward policy has been shown in Table \ref{tab:Case3}.

\section{Results and Discussion}

The analysis encompasses four key dimensions: cumulative reward, episode length, policy loss, and value prediction accuracy, tracked over 200,000 training steps.
\subsection{Performance Analysis Across Training Cases}
\subsubsection{Cumulative Reward Progression}

\begin{figure}[t]
        \subfigure[]{\includegraphics[width=0.233\textwidth]{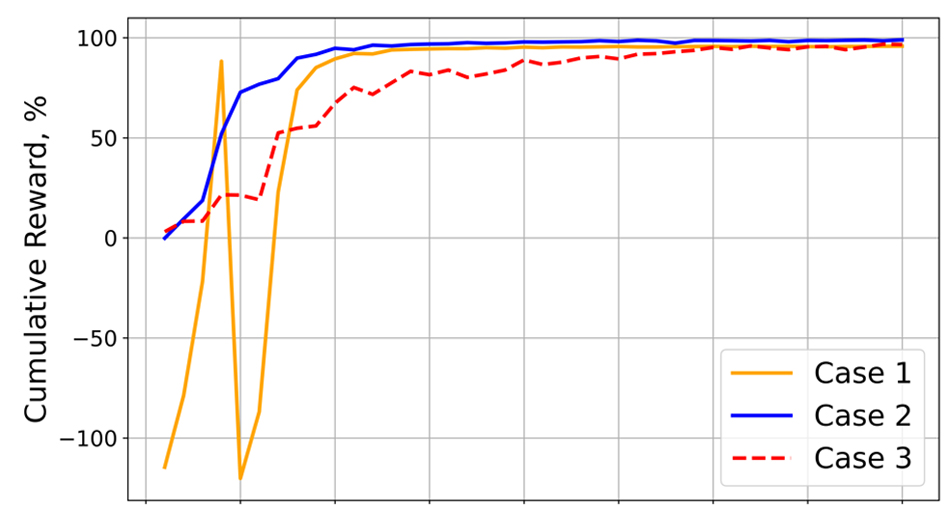}
        \label{fig:graph1}}
        \subfigure[]{\includegraphics[width=0.233\textwidth]{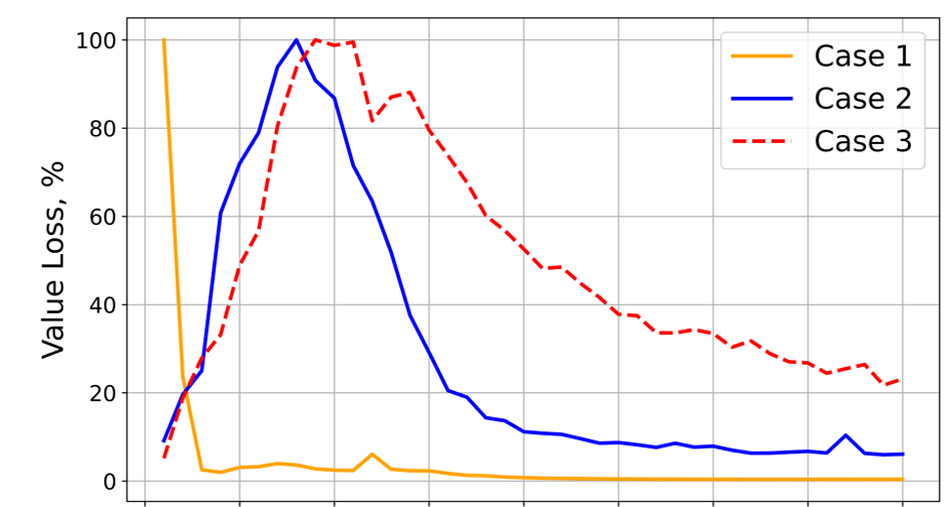}
        \label{fig:graph2}}
        \subfigure[]{\includegraphics[width=0.233\textwidth]{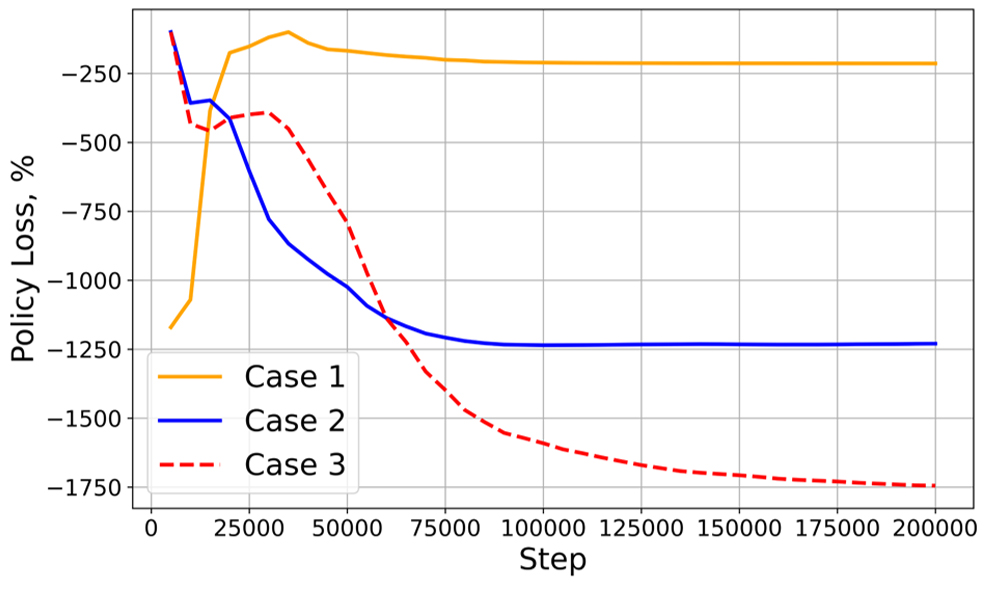}
        \label{fig:graph3}}
        \subfigure[]{\includegraphics[width=0.233\textwidth]{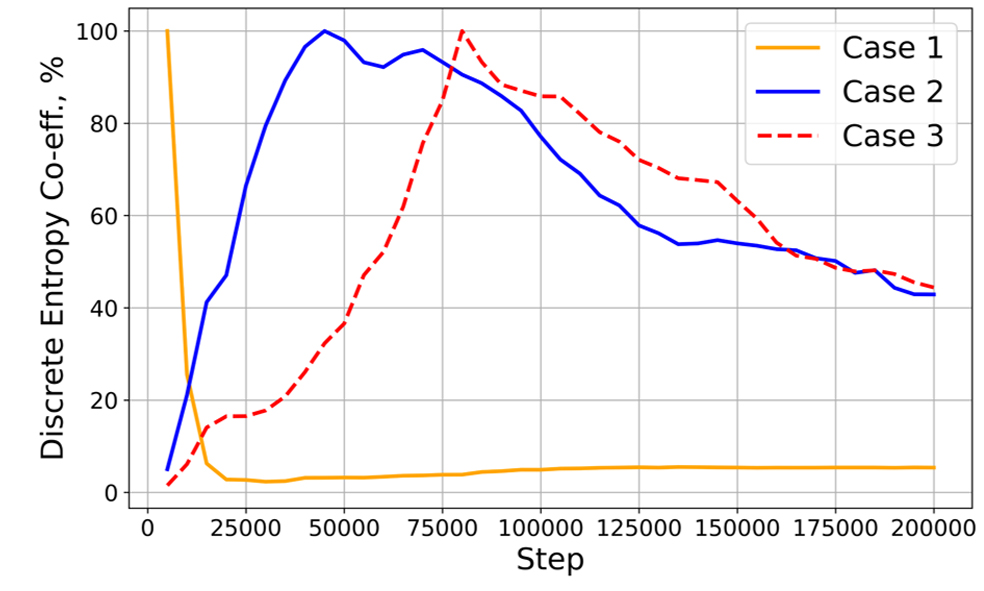}
        \label{fig:graph4}}
    \caption{Visual presentation of (a) Cumulative Reward Percentage vs Step, (b) Value Loss Percentage vs Step, (c) Policy Loss Percentage vs Step and (d) Discrete Entropy Co-eff. Percentage vs Step.}
    \label{fig:Graphs} 
\end{figure}

Figure \ref{fig:graph1} reveals distinct learning patterns across three experimental cases. In Case 1 (static target-reaching), the agent initially encountered a significant local minimum, evidenced by the sharp dip in reward percentage around the 25,000-step mark. However, the implementation of our hierarchical reward structure enabled recovery, leading to stable convergence at approximately step of 60,000. Case 3 without the hierarchical transfer training showed a gradual reward increase but with slow convergence until the step of 150,000. 
Case 2, leveraging transfer learning from Case 1, demonstrated remarkably faster convergence despite handling the more complex task of following a moving target. 
Additionally, Case 2 reached a higher cumulative reward more quickly and maintained it consistently, achieving the task's goals more effectively than others.

\subsubsection{Value Loss and Policy Optimization}
The value loss trajectories in Figure \ref{fig:graph2} provide insight into the agent's reward prediction accuracy. Case 1 exhibits rapid stabilization at approximately 5\% value loss, indicating efficient learning of the state-value function due to the relative easier task of object touching. Case 2 shows gradually increasing of value loss and then decreases rapidly to achieve stability at 100,000 steps. In Case 3, it shows a similar trend in the beginning as Case 2, but with a much slower decrement. The peak at around step 50,000 is shown for both Cases 2 and 3, likely indicating more significant adjustments during early training, which gradually stabilize, with Case 2 stabilizing at a higher speed than Case 3. Furthermore, Case 2 stabilized at a lower value loss compared to Case 3, indicating that its value function approximations are more accurate, which helps improve decision-making.

\subsubsection{Policy Evolution and Entropy Analysis}
Policy loss trends in Figure \ref{fig:graph3} reveal the adaptation characteristics of the SAC algorithm. It illustrates the tendency of the agent to change its setup actions per iteration. Case 1's policy loss stabilized at -200\%, while Case 2 showed a gradual decline to -1200\%, reflecting increasingly deterministic policy selection. Figure \ref{fig:graph3} further revealed that Cases 2 and 3 experienced greater policy adjustments, as seen by their large negative values, while Case 1 remains relatively stable. Although both cases show substantial negative policy loss, Case 2's loss stabilizes faster, implying that it found an effective policy earlier and with less overcorrection than Case 3. The discrete entropy coefficient from Figure \ref{fig:graph4} further supports this observation, demonstrating the agent's transition from exploration to exploitation, particularly evident in Case 2's rapid entropy reduction after 50,000 steps. 

\subsection{Real-world Validation Through Digital Twin Synchronization for Robotic AM Line Scanning}

\begin{figure}[t]
    \centering
    \includegraphics[width=0.475\textwidth]{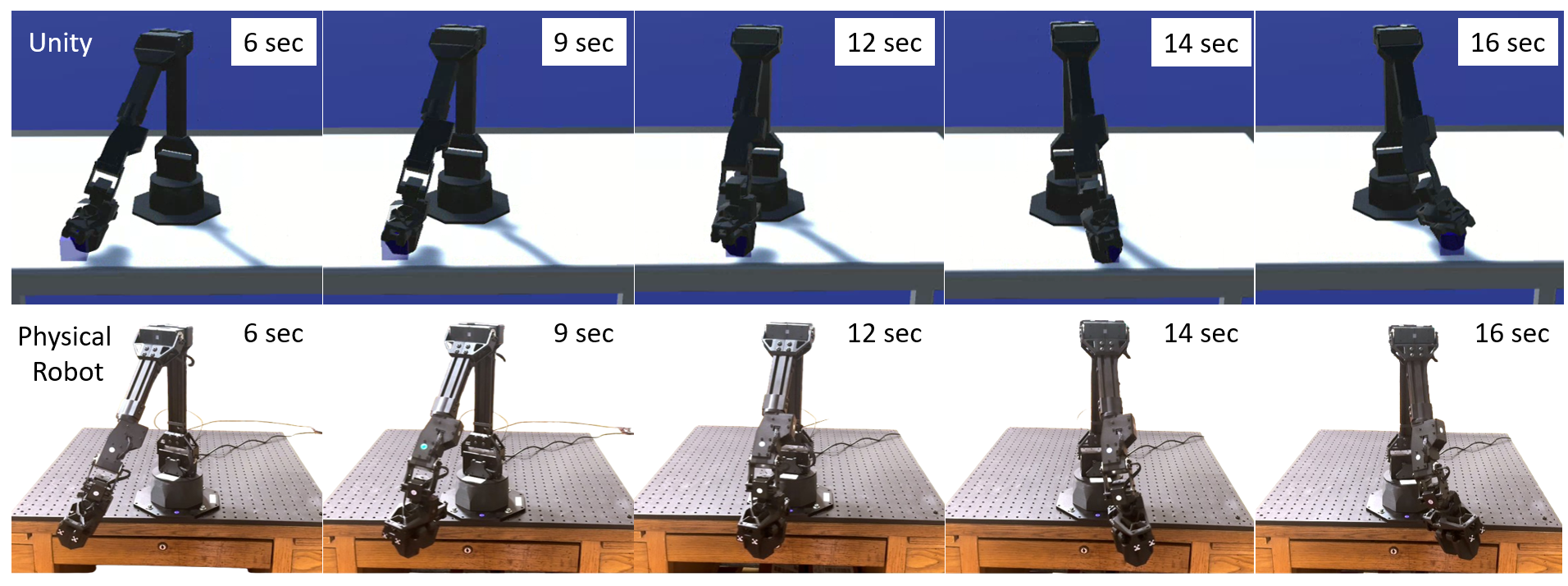}
    \caption{Validation of Digital Twin Synchronization of virtual robot arm RL environment and real-time control of physical robot AM line scanning.}
    \label{fig:SimVsReal}
\end{figure}

The practical validation of our approach is demonstrated in Figure \ref{fig:SimVsReal}, which shows the synchronized movement of virtual and physical Viper X300s robots during a linear scanning task. The temporal sequence captures key positions at 6, 9, 12, 14, and 16 seconds, revealing highly coherent behavior between the simulated and physical systems, and the observed latency remained consistently around 20 milliseconds.

Real-world validation demonstrates that the learned policies transfer smoothly and reliably from simulation to physical hardware, confirming the effectiveness of the training approach. This smooth transfer underscores the model's robustness in handling real-world complexities and suggests that simulation-based training can substantially reduce the need for extensive, time-consuming testing on physical systems. 

\section{Conclusion}

This research demonstrates the integration of RL with digital twin technology to enhance robotic control within manufacturing applications. By utilizing the SAC algorithm and the Viper X300s platform, we tackle several key challenges in smart manufacturing, ultimately developing adaptive and efficient control systems. The findings from this work highlight the transformative potential of RL-driven digital twins, showcasing their ability to enable robust, real-time synchronization between virtual simulations and physical additive manufacturing process. This breakthrough opens new avenues for optimizing automation processes and bridging the gap between simulation and real-world applications.

\section*{Acknowledgements}

The work is supported by the U.S. National Science Foundation (NSF) Foundational Research in Robotics (FRR) Award 2348013, the NSF Award 2119688, and Louisiana Board of Regents Support Fund (BoRSF) Research Competitiveness Subprogram LEQSF(2024-27)-RD-A-32.

\bibliography{references}
\bibliographystyle{IEEEtran}


\end{document}